\documentclass{article}
\usepackage{spconf,amsmath,graphicx}
\usepackage{booktabs}
\usepackage{url}
\usepackage{float}

\usepackage{enumitem}
\setlist{nosep, leftmargin=14pt}

\usepackage{mwe} 

\usepackage{amssymb}
\usepackage{hyperref}
\usepackage{booktabs}
\usepackage{array} 
\usepackage{cleveref}
\usepackage{svg}

\title{CellPilot: A unified approach to automatic and interactive segmentation in histopathology}
\name{Philipp Endres \textsuperscript{1,2,*} \thanks{\textsuperscript{*} contributed equally to this work}, Valentin Koch \textsuperscript{1,2,*},  Julia A. Schnabel \textsuperscript{1,2,3}}
\secondlinename{\textit{Carsten Marr \textsuperscript{2,$\dagger$}} \thanks{\textsuperscript{$\dagger$} corresponding author, \href{mailto:carsten.marr@helmholtz-munich.de}{carsten.marr@helmholtz-munich.de}}}

\address{\textsuperscript{1} School of Computation and Information Technology, Technical University of Munich, Germany \\
\textsuperscript{2} Helmholtz Zentrum München - German Research Center for Environmental Health, Germany \\
\textsuperscript{3} School of Biomedical Engineering and Imaging Sciences, King’s College London, UK}

\begin{document}

\maketitle
\begin{abstract}
Histopathology, the microscopic study of diseased tissue, is increasingly digitized, enabling improved visualization and streamlined workflows. An important task in histopathology is the segmentation of cells and glands, essential for determining shape and frequencies that can serve as indicators of disease. Deep learning tools are widely used in histopathology. However, variability in tissue appearance and cell morphology presents challenges for achieving reliable segmentation, often requiring manual correction to improve accuracy.
This work introduces CellPilot, a framework that bridges the gap between automatic and interactive segmentation by providing initial automatic segmentation as well as guided interactive refinement.
Our model was trained on over 675,000 masks of nine diverse cell and gland segmentation datasets, spanning 16 organs. CellPilot demonstrates superior performance compared to other interactive tools on three held-out histopathological datasets while enabling automatic segmentation.
\noindent We make the model and a graphical user interface designed to assist practitioners in creating large-scale annotated datasets available as open-source, fostering the development of more robust and generalized diagnostic models.
\footnote{\href{https://github.com/philippendres/CellPilot}{https://github.com/philippendres/CellPilot}}
\end{abstract}
\begin{keywords}
cell segmentation, medical imaging, gland segmentation
\end{keywords}
\begin{figure*}[htbp]
    \centering
    \includegraphics[width=1.0\linewidth]{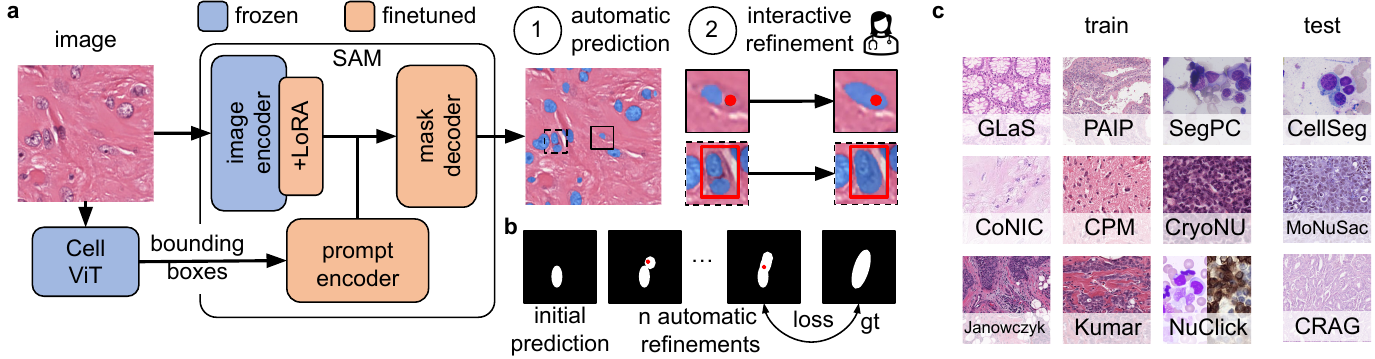}
    \caption{\textbf{CellPilot allows interactive refinement of automated predictions.} We fine-tune SAM (a) with simulated automated interactive refinements (b) on a total of nine diverse datasets spanning 16 organs and over 675,000 masks of cells and glands and test it on three held-out test sets (c). During inference, the framework works in two stages: In the first part, CellViT \cite{cellvit} is used to provide bounding boxes, which are used as input to the prompt encoder to produce automatic predictions. In the interactive stage, the user provides points or bounding boxes to add new masks or refine existing masks using positive (enlarging segmentation area) or negative prompts (decreasing it).}
    \label{fig:fig1}
\end{figure*}
\section{Introduction}
\label{sec:intro}
Histopathology, which involves the microscopic examination of diseased tissue, is essential for diagnosing cancer. The segmentation of cells and glands in tissue allows for the analysis of shape and frequency, which can be indicators of disease.  While deep learning models can automate aspects of this task, segmentation is challenging due to the variability in tissue appearance, cell morphology, and imaging conditions. Fully automatic AI-based segmentations may lack precision, while interactive approaches can be time-consuming, making a balance between automation and user control necessary.

Interactive segmentation models include the Segment Anything Model (SAM), which has been trained on a large dataset of natural images and allows using prompts like text, points, and bounding boxes to interactively segment objects \cite{SAM}. However, SAM’s performance in specialized areas like histopathology remains limited \cite{deng2023segmentmodelsamdigital}. Several approaches have been proposed to adapt SAM for medical image segmentation: Ma et al.\ fine-tune SAM’s mask decoder for medical applications (MedSAM) \cite{Ma_2024}, Zhang et al.\ apply Low-Rank Adaptation (LoRA) to the image encoder (SAMed) for fully automatic segmentation \cite{zhang2023customizedsegmentmodelmedical}.
In histopathology, SAM has been adapted by retraining specific components to enhance its performance in semantic segmentation tasks \cite{zhang2023sampathsegmentmodelsemantic, kim2023evaluationimprovementsegmentmodel}.
However, these approaches do not allow for interactive refinement \cite{zhang2023sampathsegmentmodelsemantic} or only support interactive segmentation of tumor regions \cite{kim2023evaluationimprovementsegmentmodel}.
In this work, we propose a model that bridges the gap between purely automatic segmentation and fully interactive ones. CellPilot enables both automatic segmentation and interactive refinement of cells and glands through bounding boxes and points. It is trained using nine diverse histopathology segmentation datasets, comprising a total of 7,913 images with over 675,000 masks.
To assess its performance, it is compared to the interactive methods SAM \cite{SAM}, MedSAM \cite{Ma_2024}, and SimpleClick \cite{Liu_2023_ICCV} on three held-out test sets.
Our main contributions are:
\begin{itemize}
    \item Bridging automatic and interactive segmentation: CellPilot combines the strengths of both automatic segmentation and interactive refinement, enabling users to achieve high initial accuracy while being able to correct mistakes interactively. 
    \item Large-scale, diverse training: We train our model on nine diverse datasets spanning 16 organs and over 675,000 masks, enhancing its ability to generalize across different histopathological images and conditions. 
    \item Open-access tool for dataset creation: We provide an open-source tool designed to accelerate the creation of new annotated datasets, empowering researchers and practitioners to develop fully automated models more efficiently.

\end{itemize}

\section{Experiments \& Methods}
\label{sec:experiments}

\subsection{Method}
The CellPilot framework consists of a combination of a finetuned version of SAM and CellViT (figure \ref{fig:fig1}a). 
CellViT \cite{cellvit} generates bounding boxes for cells, which are used as initial prompts for SAM to automatically segment cells. For glands, the user needs to give a prompt. The mask decoder decodes the prompts into segmentation masks for the initial automatic segmentation stage. To correct faulty automatic segmentations, a user can add positive or negative point or box prompts, adding or removing masks. 

\textbf{Training.}
We fine-tune SAM on images from histopathology by training the prompt encoder and the mask decoder and applying low-rank adaptation (LoRA) \cite{hu2021loralowrankadaptationlarge} to the image encoder.
We simulate interactive prompting during training following SimpleClick \cite{Liu_2023_ICCV}:
A random point in the ground truth mask or a box around it is randomly sampled and used as a prompt to get an initial prediction. Then, another point from the largest error region of this prediction is sampled. This can be either a positive prompt in the case of underprediction or a negative prompt in the case of overprediction. This process is iteratively repeated up to $n$ times until we get a final prediction. During training, $n \in [0,5]$ was randomly sampled. The resulting prediction is then compared with the ground truth. As the loss function, the unweighted sum of dice and cross entropy loss is used.
The CellPilot model was trained with the schedulefree AdamW optimizer \cite{defazio2024roadscheduled}. A batch size of four was used, and gradients were accumulated over four steps. The model was trained for 20 epochs, taking six days on an NVIDIA A40 GPU (48GB). A learning rate of $10^{-5}$ was used. For every instance, ten point prompts and ten box prompts were sampled to accelerate training. The ViT-B image encoder of SAM \cite{SAM} and a LoRA rank of four were used. As augmentations, RandomResizedCrops, HueSaturationValue, and D4 from the albumentation library were used \cite{info11020125}.

\textbf{Inference.}
For cells, CellViT \cite{cellvit} is used to generate initial bounding boxes and feed them into the prompt encoder of SAM. For glands, the user needs to input the prompt themselves. SAM then generates a segmentation mask for each bounding box and point prompt. The user can refine these via additional points and boxes per mask.
\begin{table}[t]
  \centering
  \begin{tabular}{lcc}
    \toprule
      dataset  & \# images & \# masks\\
    \midrule
        CoNIC \cite{graham2021coniccolonnucleiidentification}                  & 4,841 & 570,144 \\
        NuClick \cite{koohbanani2020nuclickdeeplearningframework}                 & 2,075 & 18,412 \\
        SegPC \cite{GUPTA2023102677}                     & 497 & 11,865 \\
        GlaS \cite{Sirinukunwattana2017Jan}              & 165 & 1,553 \\
        Janowczyk \cite{BibEntry2017Jul}                 & 143 & 12,726 \\
        CPM \cite{Vu2019}                                & 79 & 10,478 \\
        PAIP2023 \cite{BibEntry2024Aug}                  & 53 & 25,067 \\
        CryoNuSeg \cite{MAHBOD2021104349}                & 30 & 8,116 \\
        Kumar \cite{Kumar2017Jul}                        & 30 & 16,963 \\
    \midrule
    \midrule
        CellSeg \cite{NeurIPS-CellSeg}                  & 1,551 & 269,267 \\
        MoNuSAC \cite{Verma2021Dec}                     & 310 & 42,828 \\
        CRAG \cite{Graham2019Feb}                       & 213 & 3,053 \\

  \end{tabular}
    \caption{Train (top) and test (bottom) datasets, totaling over 9,900 images and 990,000 masks.}
    \label{tab:datasets}

\end{table}

\subsection{Datasets}
Overall, twelve datasets are used in this study (figure \ref{fig:fig1}c), with nine being in the training set and three in the held-out test set. The corresponding number of images and masks are shown in table \ref{tab:datasets}. The following datasets were used for training:
\begin{itemize}
    \item \textbf{CoNIC}: Colon nuclei segmentation dataset \cite{graham2021coniccolonnucleiidentification}.
    \item \textbf{NuClick}: Cell segmentation dataset with images of white blood cells and immunohistochemistry lymphocytes \cite{koohbanani2020nuclickdeeplearningframework}.
    \item \textbf{SegPC}: Cell segmentation dataset of multiple myeloma plasma cells obtained from bone marrow aspirate slides \cite{GUPTA2023102677}.
    \item \textbf{GlaS}: Gland segmentation dataset with images of stage T3 and T4 colorectal adenocarcinoma \cite{Sirinukunwattana2017Jan}.
    \item \textbf{Janowczyk}: Breast cancer nuclei segmentation dataset \cite{BibEntry2017Jul}.
    \item \textbf{CPM}: Cell segmentation dataset with images from patients with non-small cell lung cancer, head and neck squamous cell carcinoma, glioblastoma multiforme, and lower-grade glioma tumors \cite{Vu2019}.
    \item \textbf{PAIP2023}: Nuclei segmentation dataset with images from pancreatic and colon cancer \cite{BibEntry2024Aug}.
    \item \textbf{CryoNuSeg}: Nuclei segmentation dataset with images from ten human organs obtained via cryosection \cite{MAHBOD2021104349}.
    \item \textbf{Kumar}: Nuclei segmentation dataset featuring images from seven different types of cancer \cite{Kumar2017Jul}.

\end{itemize}

\noindent The following datasets were used for testing:
\begin{itemize}
    \item \textbf{CellSeg}: Multimodality microscopy cell segmentation dataset that includes tissue cells and cultured cells with different stains from different microscopes \cite{NeurIPS-CellSeg}.
    \item \textbf{MoNuSAC}: Nuclei segmentation dataset of four types of cancer \cite{Verma2021Dec}.
    \item \textbf{CRAG}: Colorectal adenocarcinoma gland segmentation dataset \cite{Graham2019Feb}.
\end{itemize}

\subsection{Experimental setup}
We compare CellPilot on the tasks of interactive cell and gland segmentation. Images were rescaled such that the longer side had a length of 1024 while keeping the aspect ratio and then padded to a size of 1024 $\times$ 1024 if needed. For SimpleClick, images are rescaled to a size of 448 $\times$ 448.
For each image, ten binary segmentation masks of single cells or glands were randomly chosen from the ground truth. Box prompts for each mask were sampled with a random additional margin size of zero to ten pixels around the ground truth masks. For point prompts, a random point inside the ground truth mask was used.
To simulate the interactive use of the models, we follow  Liu et al. \cite{Liu_2023_ICCV}, iteratively sampling positive or negative point prompts from the largest error region. 
\begin{figure}[H]
    \centering
    \includegraphics[width=1.0\linewidth]{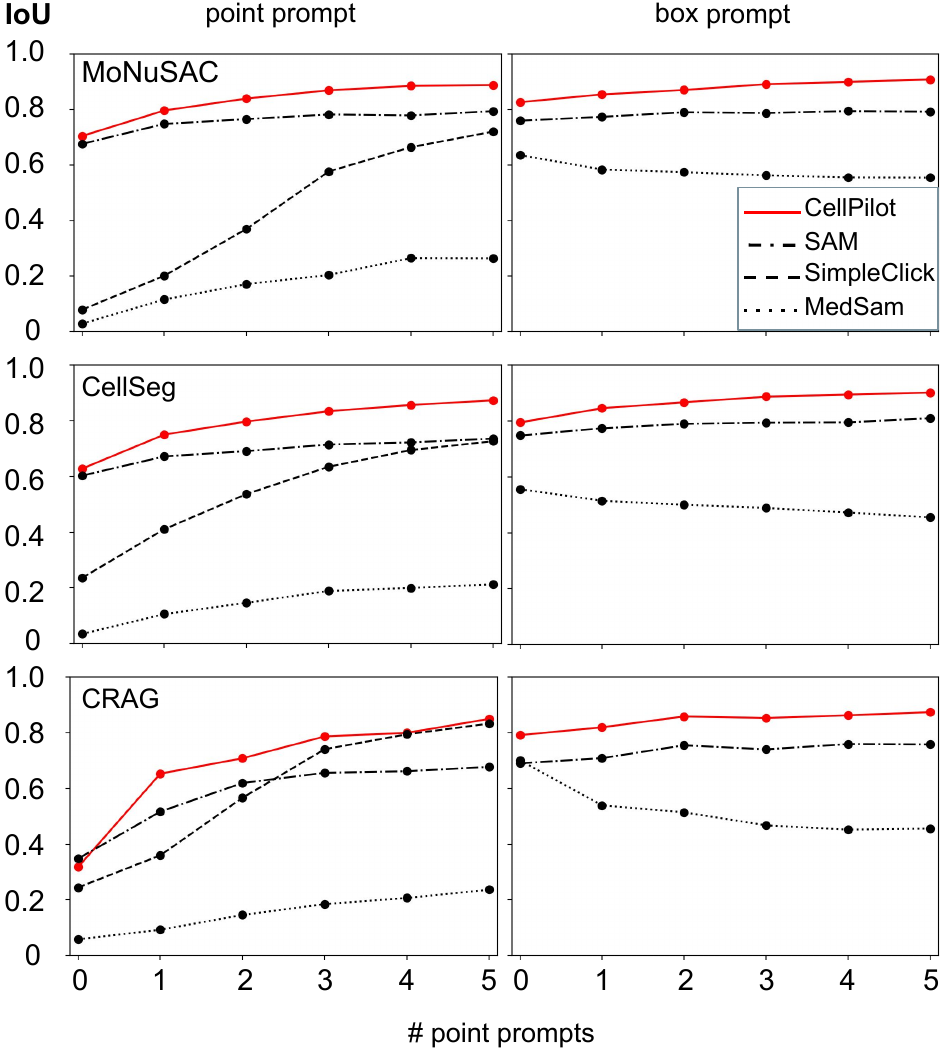}
    \caption{CellPilot outperforms MedSAM, SimpleClick and SAM on the cell segmentation datasets MoNuSAC and CellSeg, as well as on the gland segmentation dataset CRAG. We compare performance when starting with a point prompt (left) and box prompt (right) and, step by step, automatically adding points in regions of wrong prediction, simulating interactive refinement. Note that SimpleClick does not support box prompts.}
    \label{fig:fig2}
\end{figure}
\begin{figure*}[htbp]
    \centering
    \includegraphics[width=1.0\linewidth]{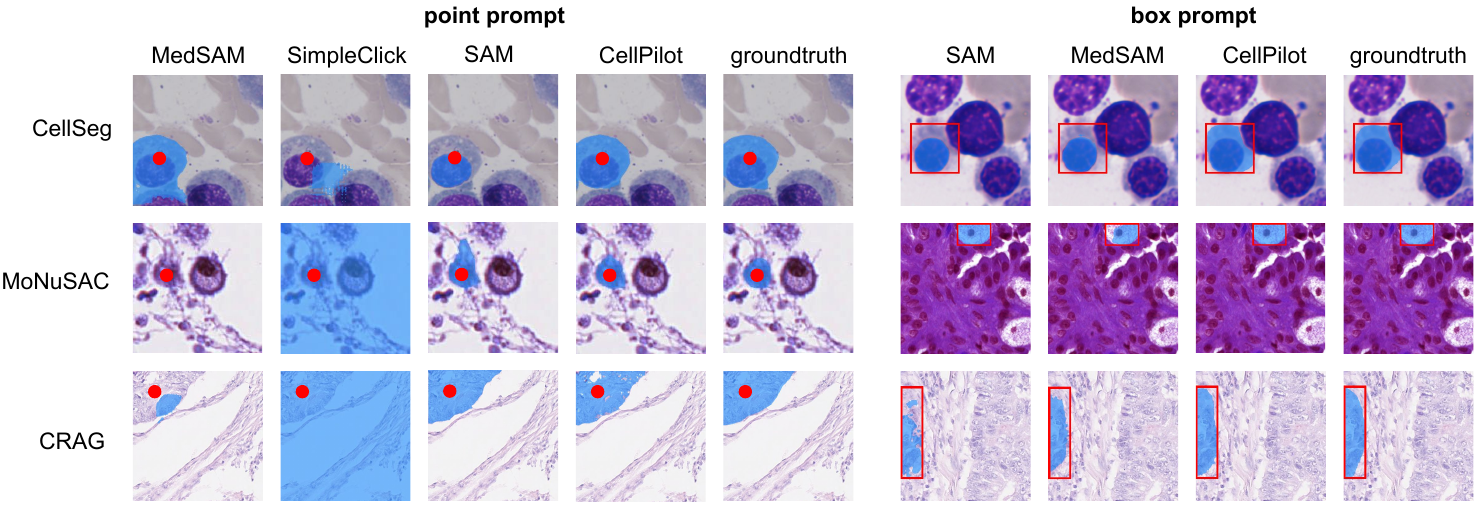}
    \caption{\textbf{CellPilot produces coherent masks.} Qualitative comparison of samples from  CellSeg, MoNuSAC, and CRAG datasets using point prompts (left) or box prompts (right). }
    \label{fig:fig3}
\end{figure*}

 \begin{table*}[t] 
    \centering 
        \begin{tabular}{lcccccc} 
        \toprule & \multicolumn{2}{c}
        {CellSeg \cite{NeurIPS-CellSeg}} & \multicolumn{2}{c}{MoNuSAC \cite{Verma2021Dec}} 
        & \multicolumn{2}{c}{CRAG \cite{Graham2019Feb}} \\ 
        & point & box & point & box & point & box \\ 
        \midrule SAM \cite{SAM} 
            & $\underline{0.60} {\scriptstyle \pm 0.31}$ 
            & $\underline{0.75} {\scriptstyle \pm 0.18}$ 
            & $\underline{0.68} {\scriptstyle \pm 0.23}$ 
            & $\underline{0.76} {\scriptstyle \pm 0.13}$ 
            & $\mathbf{0.35} {\scriptstyle \pm 0.33}$ 
            & $0.69 {\scriptstyle \pm 0.22}$ \\ 
        MedSAM \cite{Ma_2024} 
            & $0.03 {\scriptstyle \pm 0.08}$ 
            & $0.56 {\scriptstyle \pm 0.21}$ 
            & $0.03 {\scriptstyle \pm 0.08}$ & $0.64 {\scriptstyle \pm 0.19}$ & $0.06 {\scriptstyle \pm 0.11}$ 
            & $\underline{0.70} {\scriptstyle \pm 0.18}$ \\ 
        SimpleClick \cite{Liu_2023_ICCV} 
            & $0.24 {\scriptstyle \pm 0.31}$ 
            & $-$ 
            & $0.08 {\scriptstyle \pm 0.20}$ 
            & $-$ 
            & $0.24 {\scriptstyle \pm 0.30}$ 
            & $-$ \\ 
        \midrule CellPilot 
        & $\mathbf{0.63} {\scriptstyle \pm 0.27}$ 
        & $\mathbf{0.79} {\scriptstyle \pm 0.15}$ 
        & $\mathbf{0.70} {\scriptstyle \pm 0.23}$ 
        & $\mathbf{0.83} {\scriptstyle \pm 0.09}$ 
        & $\underline{0.32} {\scriptstyle \pm 0.33}$ 
        & $\mathbf{0.79} {\scriptstyle \pm 0.17}$ \\ \bottomrule    \end{tabular} 
  \caption{CellPilot achieves high mean IoU scores using a single point or box prompt in cell segmentation tasks (CellSeg, MoNuSAC). In gland segmentation (CRAG) it is closely beaten in point prompts by SAM.}
  \label{tab:combined-results}
\end{table*}
\subsection{Results}

\textbf{Interactive refinement.}
For both initially providing a point and initially providing a box prompt, CellPilot consistently improves with added corrections, showing better interactive performances than MedSAM, SimpleClick, and SAM. This can be observed across the two cell and the one gland segmentation dataset (figure \ref{fig:fig2}). MedSAM has the lowest performance; additional point prompts only help for initial point prompts, and interactively adding points even worsens the score when initially a bounding box is provided. SimpleClicks improvement is rather large when provided with additional point prompts, but it starts with comparatively low mean IoU and always performs worse than CellPilot. SAM performs second best, and generally, adding points helps, but the improvements are smaller compared to CellPilot, scoring overall worse in all evaluations except single point prompt in the gland segmentation task CRAG.

\textbf{Single prompt.}
As seen in table \ref{tab:combined-results}, the CellPilot model outperforms other state-of-the-art interactive models using point or box prompts for cell segmentation. SAM \cite{SAM} performs second best overall, while MedSAM\cite{Ma_2024}, finetuned only for bounding box prompts, is comparatively weak. SimpleClick \cite{Liu_2023_ICCV}, being trained on natural images, performs not well in cell segmentation. \\
Similarly, the CellPilot model outperforms SAM \cite{SAM}, MedSAM \cite{Ma_2024}, and SimpleClick \cite{Liu_2023_ICCV} on the task of interactive gland segmentation when a bounding box is provided. 
Using a point prompt, all models perform weak, with SAM having a slight advantage over CellPilot. \\
In figure \ref{fig:fig3}, it can be clearly seen that MedSAM cannot handle point prompts well. SAM and SimpleClick tend more towards segmenting structures that look semantically similar but are not necessarily biologically meaningful, while CellPilot segments cells and glands accurately.

\section{Conclusion}
\label{sec:conclusion}
In this work, we presented a method and an extensively trained model allowing interactive and automatic segmentation in histopathological images. Experimental results demonstrate that our interactive segmentation stage surpasses SAM, MedSAM, and SimpleClick across three different held-out datasets, performing well with both single and multiple prompts.
Despite its strengths, our model faces certain limitations. One of the key challenges lies in its reliance on minimal user input, which can lead to difficulty in distinguishing between cells and glands. Future research could explore ways to enhance our model’s ability to distinguish between cells and glands through fixed prompts or improved training techniques. CellPilot’s hybrid structure, integrating components from SAM \cite{SAM} and CellViT \cite{cellvit}, results in a larger model size. This, combined with the need to compute image embeddings twice, introduces computational overhead and can slow down the segmentation process.
Additionally, incorporating a classification module could enable semantic segmentation, allowing for more detailed identification of cell types and better distinction between healthy and abnormal tissue.

\section{Ethics statement}
\label{sec:ethics}

This research study was conducted retrospectively using human subject data made available in open access. Ethical approval was not required as confirmed by the license attached with the open access data.

\section{Acknowledgments}
\label{sec:acknowledgments}
We thank Sophia Wagner for her valuable feedback. V.K. was supported by the Helmholtz Association under the joint research school “Munich School for Data Science - MUDS”. This work was also supported by the BMBF-funded de.NBI Cloud within the German Network for Bioinformatics Infrastructure (de.NBI) (031A532B, 031A533A, 031A533B, 031A534A, 031A535A, 031A537A, 031A537B, 031A537C, 031A537D, 031A538A). C.M. has received funding from the European Research Council under the European Union’s Horizon 2020 research and innovation program (grant agreement number 866411) and is supported by the Hightech Agenda Bayern.

\bibliographystyle{IEEEbib}
\bibliography{strings,refs}

\begin{thebibliography}{10}

\bibitem{cellvit}
Fabian Hörst et~al.,
\newblock ``Cellvit: Vision transformers for precise cell segmentation and
  classification,'' 2023.

\bibitem{SAM}
Alexander Kirillov et~al.,
\newblock ``Segment anything,'' 2023.

\bibitem{deng2023segmentmodelsamdigital}
Ruining Deng et~al.,
\newblock ``Segment anything model (sam) for digital pathology: Assess
  zero-shot segmentation on whole slide imaging,'' 2023.

\bibitem{Ma_2024}
Jun Ma, Yuting He, Feifei Li, Lin Han, Chenyu You, and Bo~Wang,
\newblock ``Segment anything in medical images,''
\newblock {\em Nature Communications}, vol. 15, no. 1, Jan. 2024.

\bibitem{zhang2023customizedsegmentmodelmedical}
Kaidong Zhang and Dong Liu,
\newblock ``Customized segment anything model for medical image segmentation,''
  2023.

\bibitem{zhang2023sampathsegmentmodelsemantic}
Jingwei Zhang et~al.,
\newblock ``Sam-path: A segment anything model for semantic segmentation in
  digital pathology,'' 2023.

\bibitem{kim2023evaluationimprovementsegmentmodel}
SeungKyu Kim, Hyun-Jic Oh, Seonghui Min, and Won-Ki Jeong,
\newblock ``Evaluation and improvement of segment anything model for
  interactive histopathology image segmentation,'' 2023.

\bibitem{Liu_2023_ICCV}
Qin Liu, Zhenlin Xu, Gedas Bertasius, and Marc Niethammer,
\newblock ``Simpleclick: Interactive image segmentation with simple vision
  transformers,''
\newblock in {\em Proceedings of the IEEE/CVF International Conference on
  Computer Vision (ICCV)}, October 2023, pp. 22290--22300.

\bibitem{hu2021loralowrankadaptationlarge}
Edward~J. Hu et~al.,
\newblock ``Lora: Low-rank adaptation of large language models,'' 2021.

\bibitem{defazio2024roadscheduled}
Aaron Defazio, Xingyu~Alice Yang, Harsh Mehta, Konstantin Mishchenko, Ahmed
  Khaled, and Ashok Cutkosky,
\newblock ``The road less scheduled,'' 2024.

\bibitem{info11020125}
Alexander Buslaev, Vladimir~I. Iglovikov, Eugene Khvedchenya, Alex Parinov,
  Mikhail Druzhinin, and Alexandr~A. Kalinin,
\newblock ``Albumentations: Fast and flexible image augmentations,''
\newblock {\em Information}, vol. 11, no. 2, 2020.

\bibitem{graham2021coniccolonnucleiidentification}
Simon Graham et~al.,
\newblock ``Conic: Colon nuclei identification and counting challenge 2022,''
  2021.

\bibitem{koohbanani2020nuclickdeeplearningframework}
Navid~Alemi Koohbanani, Mostafa Jahanifar, Neda~Zamani Tajadin, and Nasir
  Rajpoot,
\newblock ``Nuclick: A deep learning framework for interactive segmentation of
  microscopy images,'' 2020.

\bibitem{GUPTA2023102677}
Anubha Gupta et~al.,
\newblock ``Segpc-2021: A challenge \& dataset on segmentation of multiple
  myeloma plasma cells from microscopic images,''
\newblock {\em Medical Image Analysis}, vol. 83, pp. 102677, 2023.

\bibitem{Sirinukunwattana2017Jan}
Korsuk Sirinukunwattana et~al.,
\newblock ``{Gland segmentation in colon histology images: The glas challenge
  contest},''
\newblock {\em Med. Image Anal.}, vol. 35:489-502., Jan. 2017.

\bibitem{BibEntry2017Jul}
Andrew Janowczyk,
\newblock ``{Use Case 1: Nuclei Segmentation - Andrew Janowczyk},'' July 2017,
\newblock [Online; accessed 10. Aug. 2024].

\bibitem{Vu2019}
Quoc~Dang Vu et~al.,
\newblock ``{Methods for Segmentation and Classification of Digital Microscopy
  Tissue Images},''
\newblock {\em Front. Bioeng. Biotechnol.}, vol. 7, 2019.

\bibitem{BibEntry2024Aug}
Jinwook Choi, Kyoungbun Lee, Won-Ki Jeong, Se~Young Chun, and Youngkwan Kim,
\newblock ``{PAIP 2023: TC prediction in pancreatic and colon cancer - Grand
  Challenge},'' Aug. 2024,
\newblock [Online; accessed 12. Aug. 2024].

\bibitem{MAHBOD2021104349}
Amirreza Mahbod et~al.,
\newblock ``Cryonuseg: A dataset for nuclei instance segmentation of
  cryosectioned h\&e-stained histological images,''
\newblock {\em Computers in Biology and Medicine}, vol. 132, pp. 104349, 2021.

\bibitem{Kumar2017Jul}
Neeraj Kumar, Ruchika Verma, Sanuj Sharma, Surabhi Bhargava, Abhishek Vahadane,
  and Amit Sethi,
\newblock ``{A Dataset and a Technique for Generalized Nuclear Segmentation for
  Computational Pathology},''
\newblock {\em IEEE Trans. Med. Imaging}, vol. 36, no. 7, pp. 1550--1560, July
  2017.

\bibitem{NeurIPS-CellSeg}
Jun Ma et~al.,
\newblock ``The multi-modality cell segmentation challenge: Towards universal
  solutions,''
\newblock {\em Nature Methods}, 2024.

\bibitem{Verma2021Dec}
Ruchika Verma et~al.,
\newblock ``{MoNuSAC2020: A Multi-Organ Nuclei Segmentation and Classification
  Challenge},''
\newblock {\em IEEE Trans. Med. Imaging}, vol. 40, no. 12, pp. 3413--3423, Dec.
  2021.

\bibitem{Graham2019Feb}
Simon Graham et~al.,
\newblock ``{MILD-Net: Minimal information loss dilated network for gland
  instance segmentation in colon histology images},''
\newblock {\em Med. Image Anal.}, vol. 52:199-211., Feb. 2019.

\end{thebibliography}

\end{document}